\pdfoutput=1
\documentclass[11pt]{article}

\usepackage[]{acl}
\usepackage{times}
\usepackage{latexsym}
\usepackage{enumitem}
\usepackage{amsmath}
\usepackage{subfigure}
\usepackage{multirow}
\usepackage{booktabs}
\newcommand{\modi}{\textcolor{black}}

\usepackage[T1]{fontenc}
\usepackage[utf8]{inputenc}

\usepackage{microtype}

\usepackage{graphicx}

\title{Retrieval-Augmented Machine Translation with Unstructured Knowledge}

\author{Jiaan Wang, \ Fandong Meng\thanks{ \ \ Corresponding author.}, \ Yingxue Zhang, \ Jie Zhou \\
Pattern Recognition Center, WeChat AI, Tencent Inc \\ 
\texttt{\{torchwang,fandongmeng,yxuezhang,withtomzhou\}@tencent.com}
}

\begin{document}
\maketitle
\begin{abstract}
Retrieval-augmented generation (RAG) introduces additional information to enhance large language models (LLMs).
In machine translation (MT), previous work typically retrieves in-context examples from paired MT corpora, or domain-specific knowledge from knowledge graphs, to enhance MT models.
However, a large amount of world knowledge is organized in unstructured documents, and might not be fully paired across different languages.
In this paper, we study retrieval-augmented MT using unstructured documents.
Specifically, we build RAGtrans, the first benchmark to train and evaluate LLMs' retrieval-augmented MT ability.
\modi{RAGtrans contains 169K MT samples collected via GPT-4o and human translators.}
Besides, documents from various languages are also provided to supply the knowledge to these samples.
Based on RAGtrans, we further propose a multi-task training method to teach LLMs how to use information from multilingual documents during their translation.
The method uses existing multilingual corpora to create auxiliary training objectives without additional labeling requirements.
Extensive experiments show that the method improves LLMs by 1.6$\sim$3.1 BLEU and 1.0$\sim$2.0 COMET scores in En$\Rightarrow$Zh, \modi{and 1.7$\sim$2.9 BLEU and 2.1$\sim$2.7 COMET scores in En$\Rightarrow$De.}
We also conclude the critical difficulties that current LLMs face with this task.\footnote{\url{https://github.com/krystalan/RAGtrans}}
\end{abstract}

\section{Introduction}

Retrieval-augmented generation (RAG) has grown into a practical paradigm in the development of large language models (LLMs).
With the help of retrieved information, LLMs could generate more accurate and knowledge-enrich responses~\cite{li2022survey,gao2023retrieval}.

Previous work brought RAG into machine translation (MT), and could be mainly classified into the following two streams:
(1) \emph{Retrieving in-context examples} (also known as ``translation memory''): for a source sentence, a few studies retrieve the relevant paired sentences from bilingual corpora to enhance MT models~\cite{zhang-etal-2018-guiding,bulte-tezcan-2019-neural,khandelwalnearest,he-etal-2021-fast,hoang-etal-2023-improving}. Further, \citet{cai-etal-2021-neural} relax the bilingualism limitation, and try to directly retrieve similar target-language translations to enhance models.
(2) \emph{Retrieving knowledge triplets}: the others retrieve relevant information from knowledge graphs to let the models know domain or cultural knowledge w.r.t. the source sentences~\cite{conia-etal-2024-towards,chen2024crat}.
Despite the great success that has been achieved, a large amount of world knowledge is organized in unstructured documents, and might not be fully paired across different languages.
This unstructured knowledge is neglected by previous work.
For example, Wikipedia serves as an encyclopedia of world knowledge.
Most of its information is listed in documents.
Besides, for a piece of specific knowledge, Wikipedia does not always provide it in all languages. Though multilingual information of some general knowledge is provided, their content might be differentiated among different languages~\cite{perez-beltrachini-lapata-2021-models}.

In this paper, we study retrieval-augmented MT using unstructured documents.
Since we are the first to study this topic and previous datasets do not support the research, we first build a benchmark dataset, named \textbf{RAGtrans}.
In detail, RAGtrans is collected based on Wikipedia with three key features:
(\romannumeral1) \emph{Knowledge-intensive sentences}: RAGtrans randomly selects \modi{169K English sentences} from Wikipedia as the source sentences, which generally come from the lead paragraphs of different Wikipedia pages, containing knowledge-intensive semantics.
Thus, understanding these source sentences tends to require additional knowledge.
(\romannumeral2) \emph{Useful relevant documents}: To achieve retrieval-augmented MT, for each source sentence, its following content on the Wikipedia page (in English) could serve as its relevant document.
(\romannumeral3) \emph{Transferability to multilingual RAG}: Wikipedia also provides multilingual parallel content. Therefore, for a source sentence, its relevant knowledge in different languages can also serve as relevant documents.
As a result, MT models can leverage knowledge from multilingual documents beyond the source and the target languages.
In this work, we choose Chinese, German, French and Czech.
After collecting the source English sentences and relevant documents, we randomly split them into training, validation and testing sets.
For training and validation samples, we employ GPT-4o~\cite{hurst2024gpt} to collect the \modi{Chinese or German translation}; while we employ professional human translators to perform the same process for the testing samples.
Finally, \modi{RAGtrans involves 79K retrieval-augment En$\Rightarrow$Zh and 90K En$\Rightarrow$De MT samples}\footnote{En$\Rightarrow$Zh means the translation from English to Chinese; En$\Rightarrow$De means the translation from English to German. En: English; Zh: Chinese; De: German}, each of which contains an English source sentence, a document in English, Chinese, German, French or Czech, and the corresponding translation.

Based on RAGtrans, we train LLMs and evaluate their retrieval-augmented MT performance from the following settings:
(1) \emph{Golden evaluation}: providing LLMs with the golden relevant documents during data collection, and testing the translation performance.
(2) \emph{Robustness evaluation}: providing irrelevant documents to test the LLMs' robustness.
(3) \emph{Full Wiki evaluation}: Equipping LLMs with a (multilingual) retriever to first retrieve relevant documents from the whole Wikipedia, and then evaluate their retrieval-augmented MT ability.

Furthermore, during the application phase of a retrieval-augmented MT model, the model might receive multiple documents from various languages.
These multilingual documents are not restricted to parallel documents and can convey diverse meanings.
In light of this, we propose a multi-task training method to enhance LLMs' ability to leverage multilingual knowledge.
Specifically, we design three training objectives, \emph{i.e.}, cross-lingual information completion, self-knowledge-enhanced translation and cross-lingual relevance discrimination.
Among them, cross-lingual information completion and cross-lingual relevance discrimination train LLMs to refine and judge information from multilingual documents.
Self-knowledge-enhanced translation lets LLMs generate relevant knowledge in various languages for the source sentences, and then perform MT with the help of its multilingual self-knowledge.
The multi-task training samples of these objectives can be automatically created from existing multilingual corpora, and do not need any additional labeling costs.
Experiments on RAGtrans show that the multi-task training method improves LLMs' ability to leverage relevant knowledge.
Using Qwen2.5-7B~\cite{yang2024qwen2} as the backbone, compared with simply instruction-tuning on RAGtrans, the retrieval-augmented MT performance is improved by 1.6$\sim$3.1 BLEU and 1.0$\sim$2.0 COMET scores in En$\Rightarrow$Zh, and \modi{1.7$\sim$2.9 BLEU and 2.1$\sim$2.7 COMET scores in En$\Rightarrow$De}.
Finally, we discuss specific challenges that current approaches faced with this task and give multiple promising directions for future research.

Our main contributions are concluded as follows:
\begin{itemize}[leftmargin=*,topsep=0pt]
\setlength{\itemsep}{0pt}
\setlength{\parsep}{0pt}
\setlength{\parskip}{0pt}
% (1) 
\item To the best of our knowledge, we are the first to study retrieval-augmented MT using unstructured knowledge. To this end, we construct the first corresponding benchmark dataset, \emph{i.e.}, RAGtrans, containing 169K translation samples collected via GPT-4o and human translators.

\item We propose a multi-task training method with three designed training objectives to improve LLMs' retrieval-augmented MT ability. The multi-task training samples are low-cost, and do not require additional labeling costs. % 
\item In-depth analyses of the retrieval-augmented MT results on automatic evaluation and human evaluation provide a deeper understanding of this research direction. 

\end{itemize}

\section{RAGtrans}
\label{sec:ragtrans}
In this section, we first discuss how we select English source sentences and their relevant documents from Wikipedia (\S~\ref{subsec:2.1}).
Then, we introduce the details of the data translation via GPT-4o and human translators (\S~\ref{subsec:2.2}).
Finally, we give statistical analyses of RAGtrans (\S~\ref{subsec:2.3}), and provide the details of benchmark settings (\S~\ref{subsec:2.4}).

\subsection{Data Selection}
\label{subsec:2.1}

When deciding the source sentences we focus on, there are three requirements that should be met:
(1) The source sentences should involve knowledge-intensive semantics, otherwise, they might be trivial to translate and do not need additional knowledge.
(2) It should be convenient to collect their relevant documents from existing resources, otherwise, annotating relevant documents is labor-intensive.
(3) It should also be possible to collect relevant documents in other languages.
This is because world knowledge is recorded in multilingual form. If we restrict the language of the retrieved documents, the practicality will decrease.

% \footnote{\url{https://www.wikipedia.org/}}
After carefully comparing existing open-source resources, we decide to select both source sentences and relevant documents from Wikipedia.
Formally, we denote an English document on a Wikipedia page as $D^{\text{en}} = \{p^{\text{en}}_1, p^{\text{en}}_2, ..., p^{\text{en}}_{|D|}\}$, where $p^{\text{en}}_i$ indicates the $i$-th paragraph in $D^{\text{en}}$.
Inspired by \citet{perez-beltrachini-lapata-2021-models}, the lead paragraph of a Wikipedia page contains knowledge-intensive semantics.
Thus, we use $p^{\text{en}}_1$ from each randomly selected Wikipedia page as a source sentence to meet the requirement (1).
In view of the paragraphs on the same Wikipedia page are generally highly related, to meet the requirement (2) for $p^{\text{en}}_1$, we randomly select its consecutive paragraphs, \emph{i.e.}, $D^{\text{en}} \setminus p^{\text{en}}_1$, as its relevant document.
To further collect relevant documents beyond English, \emph{i.e.}, the requirement (3), we exploit the parallel documents in other languages of $D^{\text{en}}$ provided by Wikipedia. In this work, we choose Chinese, German, French and Czech, and denote the corresponding parallel documents as $D^{\text{zh}}$, $D^{\text{de}}$, $D^{\text{fr}}$ and $D^{\text{cs}}$, respectively.
Given this, the consecutive paragraphs from $D^{\textit{l}} \setminus p^{\textit{l}}_1 (\textit{l} \in \{\text{zh}, \text{de}, \text{fr}, \text{cs}\})$ form as the relevant document in other languages.

To ensure robustness, for a small number of source sentences, we randomly select documents from the whole Wikipedia to serve as noisy documents.
After the above process, we obtain \modi{169K English source sentences}. For each sentence, a (relevant or noisy) document in English, Chinese, German, French or Czech is also provided.

\subsection{Translation Annotation}
\label{subsec:2.2}

For a given source sentence, we next collect its translation in the target language conditioned on the corresponding document.
\modi{In this work, we focus on En$\Rightarrow$Zh and En$\Rightarrow$De translation, and we collect the Chinese translation for the 79K English sentences, while German translation for the remaining 90K English sentences.}
Considering the trade-off between quality and cost, we decide to translate the source sentences of the training and validation sets via GPT-4o~\cite{hurst2024gpt}, while those of the testing set are translated via human translators.

\begin{figure}[t]
\centerline{\includegraphics[width=0.48\textwidth]{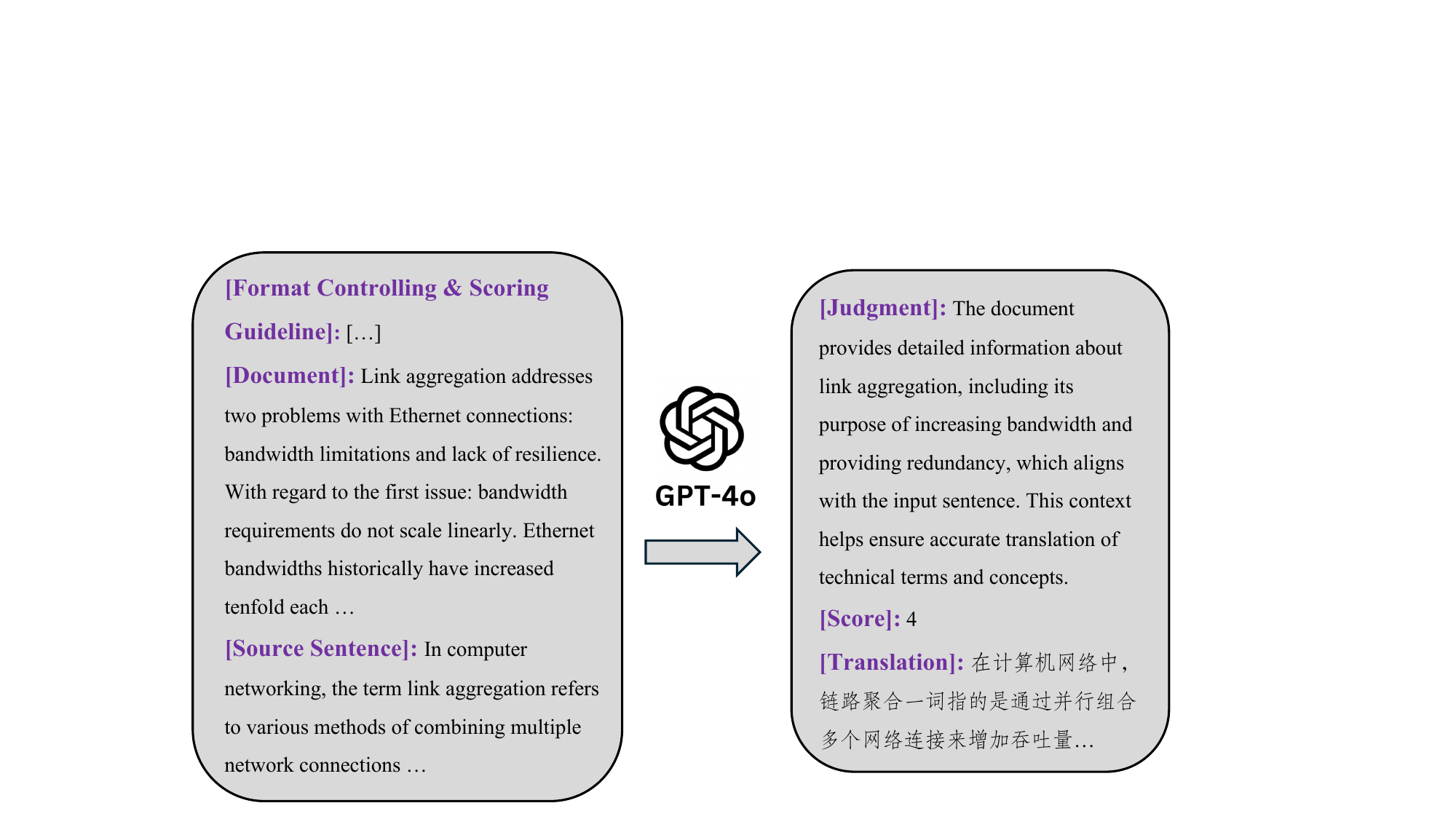}}
\caption{The overview of GPT-4o translation.}
\label{fig:gpt_4o_translation}
\end{figure} 

% \vspace{0.5ex}
\noindent \textbf{GPT-4o translation.}
Given a source sentence and the corresponding document, we prompt GPT-4o to perform retrieval-augmented MT to collect its Chinese or German translation.
To achieve better translation, we let GPT-4o first judge the relevance of the given document to respond with a judgment and a 5-point rating, and then translate the sentence in a chain-of-thought (CoT) manner.
Figure~\ref{fig:gpt_4o_translation} gives a brief overview of the process.
We also provide an example of the complete prompt, quality analysis and other details in Appendix~\ref{appendix:gpt4o_trans}.

% \vspace{0.5ex}
\noindent \textbf{Human translation.}
\modi{For source sentences in the testing set, we employ 17 professional human translators to collect the Chinese or German translations.
Among them, 10 translators are En$\Rightarrow$Zh translators while 7 translators are En$\Rightarrow$De translators.
All translators are native Chinese.
For En$\Rightarrow$Zh translators, they major in English, and have passed the English translator qualification.
For En$\Rightarrow$De translators, they major in German, and have passed the German translator qualification.}
We only provide the source English sentences to the annotators, and encourage them to search for the information they need from Wikipedia.
In addition, there are five data reviewers with rich experience in checking translation quality, and 20\% of the sentences translated by each translator are checked by a reviewer.
If the translation accuracy is lower than 95\%, the translator needs to modify all his/her translations under the guidance of the reviewer.

\modi{Finally, we obtain 79K En$\Rightarrow$Zh and 90K En$\Rightarrow$De retrieval-augmented MT samples. Among them, 77K En$\Rightarrow$Zh and 88K En$\Rightarrow$De samples from the training and validation sets are translated by the GPT-4o translator.}
Each sample can be formulated as a triplet $\langle s, d^\textit{l} ,t \rangle$, where $s$ and $t$ indicate the source English sentence and its translation, respectively.
$d^\textit{l}$ indicates the given document for $s$, and $l \in \{\text{en}, \text{zh}, \text{de}, \text{fr}, \text{cs}\}$ represents its language. 
In addition, \modi{2K En$\Rightarrow$Zh and 2K En$\Rightarrow$De samples from the testing set are translated by human translators.}
For each test sample, we provide the relevant English, Chinese and German documents (derived from the corresponding and parallel Wikipedia documents) in RAGtrans.
Thus, a testing sample could be formulated as a quintuple $\langle s, d^\text{en}, d^\text{zh}, d^\text{de},t \rangle$.

\begin{table}[t]
\centering
\resizebox{0.45\textwidth}{!}
{
\begin{tabular}{lc|ccc|ccc}
\toprule[1pt]
\multicolumn{2}{c}{Document}                     &  \multicolumn{3}{c}{En$\Rightarrow$Zh}   &  \multicolumn{3}{c}{En$\Rightarrow$De}       \\
\cmidrule(r){1-2} \cmidrule(r){3-5} \cmidrule(r){6-8} \multicolumn{1}{c}{Type} & \multicolumn{1}{c}{Lang.} & \multicolumn{1}{c}{Train} & Valid &     \multicolumn{1}{c}{Test}     & \multicolumn{1}{c}{Train} & Valid &     Test      \\ \midrule[1pt]
\multirow{5}{*}{Relevant}        & En            & 19,500      & 500        & \multirow{10}{*}{2,000}  &  19500 & 500 & \multirow{10}{*}{2,000}  \\
& Zh            & 19,500      & 500        &      & 19500 & 500 &                   \\
& De            & 9,700       & 300        &      & 19500 & 500 &                   \\
& Fr            & 9,700       & 300        &       & 9700 & 300 &                 \\
& Cs            & 9,700       & 300        &       & 9700 & 300 &                  \\ % \cmidrule[1pt]{1-4}
\multirow{5}{*}{Noisy}           & En            & 1,850       & 150        &        & 1850 & 150 &                 \\
& Zh            & 1,850       & 150        &         & 1850 & 150 &                \\
& De            & 900        & 100        &        & 1850 & 150 &                 \\
& Fr            & 900        & 100        &           & 900 & 100 &              \\
& Cs            & 900        & 100        &             & 900 & 100 &            \\ \midrule[1pt]
\multicolumn{2}{c}{Total}                        & 74,500      & 2,500       & \multicolumn{1}{c}{2,000} & \multicolumn{1}{c}{85250} & 2750 & 2000                   \\ \bottomrule[1pt]
\end{tabular}
}
\caption{\modi{The number of retrieval-augment MT samples in RAGtrans w.r.t. different types and different languages (Lang.) of documents.}} 
\label{table:data_statistics_number_of_samples}
\end{table}

\begin{figure}[t]
\centerline{\includegraphics[width=0.42\textwidth]{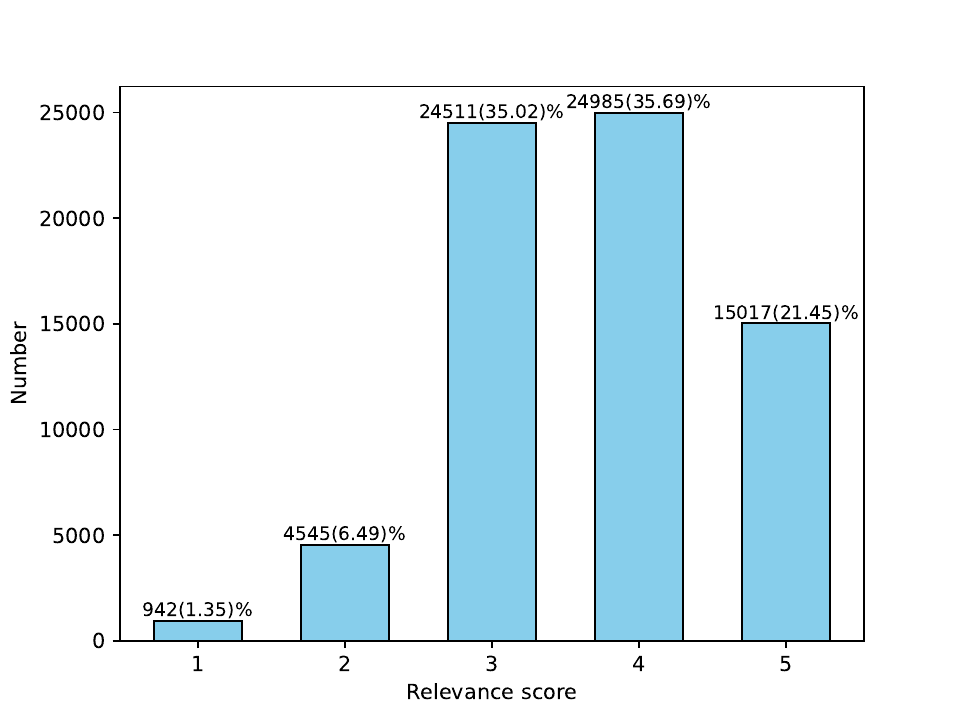}}
\caption{The distribution of relevance scores.}
\label{fig:relevance_score_distribution}
% \vspace{-0.5cm}
\end{figure}

\subsection{Data Statistics}
\label{subsec:2.3}

Table~\ref{table:data_statistics_number_of_samples} shows the number of samples w.r.t. different types (relevant or noisy) and different languages of the given documents.
In the En$\Rightarrow$Zh training and validation sets, 8.59\% and 24\% of samples are associated with noisy documents.
We emphasize the ratio of noisy documents in the validation set since robustness is vital in real applications. En$\Rightarrow$De samples also show the same tendency.
Moreover, for the training and validation samples, GPT-4o also outputs a 5-point rating (named relevance score) w.r.t. the given documents (c.f. right middle section in Figure~\ref{fig:gpt_4o_translation}).
Taking En$\Rightarrow$Zh as an example, for samples with relevant documents, we also calculate the distribution of their relevance scores.
As shown in Figure~\ref{fig:relevance_score_distribution}, more than 92\% of the documents are regarded as ``relevant'' ($\geq3$) by GPT-4o.
For samples with noisy documents, 99.93\% (6,995/7,000) of samples are judged as ``1'', while the remaining 0.07\% (5/7000) are ``2''. 

\begin{table}[t]
\centering
\resizebox{0.45\textwidth}{!}
{
\begin{tabular}{clcccc}
\toprule[1pt]
& & Min. & Max.  & Avg.   & 95th ptcl. \\ \midrule[1pt]
\multirow{7}{*}{En$\Rightarrow$Zh} & Source (En) & 5   & 526  & 85.83  & 173        \\
& Target (Zh) & 6   & 669  & 100.26 & 202        \\
& Document (En)  & 2   & 3,254 & 326.08 & 874        \\
& Document (Zh)  & 2   & 4,456 & 349.65 & 925        \\
& Document (De)  & 33  & 1,065 & 367.43 & 791        \\
& Document (Fr)  & 29  & 3,481 & 365.09 & 902        \\
& Document (Cs)  & 38  & 962  & 369.84 & 769        \\ \midrule[1pt]
\multirow{7}{*}{En$\Rightarrow$De} & Source (En) &  6 & 559  & 83.19  &  168 \\
& Target (De) & 8  &  679 & 101.27  &  206 \\
& Document (En)  &  26 & 1155  &  322.85 &  786 \\
& Document (Zh)  &  11 & 1446  &  339.36 &  854 \\
& Document (De)  &  31 & 1026  &  360.68 &  784 \\
& Document (Fr)  &  29 & 3265  &  362.31 &  888 \\
& Document (Cs)  &  41 & 943  & 369.05  &  779 \\ \bottomrule[1pt]

\end{tabular}
}
\caption{\modi{The minimum (Min.), maximum (Max.), average (Avg.) and 95th percentile (ptcl.) of tokens in the source sentence, target translation, and documents.}} 
\label{table:data_statistics_lens}
\end{table}

As for the length of source sentences, target sentences and documents, we use tiktoken\footnote{\url{https://github.com/openai/tiktoken}} to calculate their token-level length.
Table~\ref{table:data_statistics_lens} shows the minimum, maximum, and average length.
95th percentile of length is also provided.
We find that an extremely small number of documents only have single-digit tokens, which should be considered as noises, and we reserve these samples under the robustness consideration.

\subsection{Benchmark Settings}
\label{subsec:2.4}

We design three benchmark settings to evaluate the retrieval-augmented MT models:
(1) \textbf{Golden Evaluation}: For each testing sample $\langle s, d^\text{en}, d^\text{zh}, d^\text{de},t \rangle$, we give the source sentence ($s$) and a golden relevant document ($d^\text{en}$/$d^\text{zh}$/$d^\text{de}$) to the model, and evaluate models' translation.
(2) \textbf{Robustness Evaluation}: We give $s$ and an irrelevant document (randomly selected from Wikipedia) to the model, and evaluate its translation.
(3) \textbf{Full Wiki Evaluation}: This setting equips the MT models with a retriever, and truly tests models' retrieval-augmented MT ability.
For a given $s$, a retriever should first retrieve relevant documents from the whole Wikipedia, and then input both $s$ and retrieved documents to the MT model to get translation.

\section{Multi-Task Training}
\label{sec:3}

To further enhance LLMs' retrieval-augmented MT ability, we propose a multi-task training method, named \textbf{CSC}, which involves three designed training objectives, \emph{i.e.}, \textbf{C}ross-lingual information completion, \textbf{S}elf-knowledge-enhanced translation and \textbf{C}ross-lingual relevance discrimination.
In this section, we first introduce these objectives (\S~\ref{subsec:3.1}) and then discuss how to create their training samples from existing corpora (\S~\ref{subsec:3.2}).

\subsection{Multi-Task Training Objectives}
\label{subsec:3.1}

When developing a retrieval-augment MT model in real applications, it is possible to retrieve information from multilingual knowledge bases for a given source sentence. As a result, the model might receive multiple documents from various languages, extending beyond both the source and target languages. 
In such a situation, the challenge of effectively refining knowledge from these multilingual documents becomes increasingly significant.
To this end, we design three training objectives:

(1) \emph{Cross-lingual information completion.}
Given a multilingual document $d^\text{mix}$ whose paragraphs might be in different languages, and its truncated summary $\hat{y}$ in one language (\emph{e.g.}, English), we require LLMs to expand $\hat{y}$ to a complete summary $y$.
Formally, this objective can be formulated as $\Theta(y|d^\text{mix},\hat{y})$, where $\Theta$ denotes the LLMs.

% \vspace{0.5ex}
(2) \emph{Self-knowledge-enhanced translation.}
As revealed by recent RAG studies~\cite{wang2023self,liu2024ra,asai2024selfrag}, RAG models can achieve better performance with the help of their own knowledge.
Inspired by this idea, we design self-knowledge-enhanced translation.
Specifically, given a source sentence $s$, LLMs first generate its relevant document $\tilde{d}^l$ in a specific language $l\in\{\text{en}, \text{zh}, \text{de}, \text{fr}, \text{cs}\}$ and then incorporate the document to translate $s$ to $t$, denoted as $\Theta(t|\tilde{d}^{l}|s)$.

% \vspace{0.5ex}
(3) \emph{Cross-lingual relevance discrimination.}
Given that the retrieved documents may be in various languages, a crucial capability is to assess the relevance between two texts in different languages.
To this end, given a document pair $\langle d^{l_1}, d^{l_2} \rangle$ ($l_1$ $\neq$ $l_2$), $l_1$ and $l_2$ denote the languages of the documents, the model is required to generate the relevance between $d^{l_1}$ and $d^{l_2}$, denotes as $\operatorname{r}(d^{l_1}, d^{l_2})$.
The object can be formulated as $\Theta(r | d^{l_1}, d^{l_2})$

\subsection{Multi-Task Training Samples}
\label{subsec:3.2}

To create the samples for these training objectives, a principle is to reformulate existing corpora instead of labeling new data to ensure scalability.

(1) \emph{Cross-lingual information completion.}
To create the multilingual document $d^\text{mix}$ and its summary $y$, we reformulate the Wikipedia corpus. 
As revealed by~\citet{perez-beltrachini-lapata-2021-models}, the lead paragraph in a Wikipedia page could be regarded as its summary.
Given this, for an English Wikipedia page $D^\text{en}$, we extract its lead paragraph (\emph{i.e.}, $p^\text{en}_1$) as $y$, and randomly truncate $y$ to $\hat{y}$.
We next construct $d^\text{mix}$ from the remaining paragraphs $\hat{D}^\text{en} = \{p^\text{en}_i | i\geq2\}$, and the parallel counterparts in other languages, \emph{i.e.}, $\hat{D}^\text{zh}$, $\hat{D}^\text{de}$, $\hat{D}^\text{fr}$ and $\hat{D}^\text{cs}$ in this work.
Since there might be redundant information across parallel paragraphs, we use MMR algorithm~\cite{carbonell1998use} to select paragraphs from these multilingual paragraphs, \emph{i.e.}, $\bigcup_{\textit{l}} \hat{D}^\textit{l}$, to form $d^\text{mix}$.
MMR is a statistical algorithm that iteratively selects key paragraphs from the given document, at each selection step, it evaluates the relevance and redundancy of the unselected paragraphs in relation to the selected ones to determine which paragraph to select in that step.

(2) \emph{Self-knowledge-enhanced translation.}
We reformulate previous multilingual MT corpora to create samples. In detail, we use TED talk corpus~\cite{aharoni-etal-2019-massively}, where each sentence is provided with multilingual parallel sentences.
For an English sentence $s^\text{en}$, we input the sentence or its parallel sentences in other languages (\emph{i.e.}, $s^\textit{l}$) to a LLM $\Theta$, and prompt $\Theta$ to generate its relevant knowledge in the corresponding languages, \emph{i.e.}, $\tilde{d}^\textit{l}$.
In this way, $\tilde{d}^\textit{l}$ could be used as a relevant document to translate $s^\text{en}$ to other languages.

(3) \emph{Cross-lingual relevance discrimination.} We reformulate the parallel Wikipedia documents to create the samples.
Intuitively, randomly selected paragraphs from two parallel documents are relevant; while those from different documents are irrelevant.
In this way, we create the document pair and the corresponding boolean relevance.

\begin{table*}[t]
\centering
\resizebox{0.99\textwidth}{!}
{
\begin{tabular}{clc|c|c|c|c|c}
\toprule[1pt]
 & & \multicolumn{3}{c|}{English$\Rightarrow$Chinese (En$\Rightarrow$Zh)} & \multicolumn{3}{c}{English$\Rightarrow$German (En$\Rightarrow$De)} \\
 &  & \multicolumn{1}{c}{Zero-Shot LLMs}      & \multicolumn{1}{c}{SFT LLMs}    & \multicolumn{1}{c|}{SFT+CSC LLMs}  & \multicolumn{1}{c}{Zero-Shot LLMs}      & \multicolumn{1}{c}{SFT LLMs}    & \multicolumn{1}{c}{SFT+CSC LLMs} \\  \midrule[1pt]
% & & BLEU            & COMET           & GRB                 & \multicolumn{1}{c}{GRF}             & BLEU            & COMET           & GRB                 & \multicolumn{1}{c}{GRF}  & BLEU            & COMET           & GRB                 & GRF              \\ \midrule[1pt]
\multicolumn{8}{c}{\texttt{w/ Empty Document}} \\ \midrule
1 & Qwen2.5-7B  & 50.1 / 84.6 / 85.5 / 87.0    & 54.6 / 86.7 / 88.5 / 89.1    & {56.8} / {87.7} / {91.2} / {91.8}  & 56.1 / 66.7 / 74.7 / 78.1     &      59.2 / 69.7 / 81.5 / 83.4     &      {60.9} / {71.8} / {84.4} / {86.9}    \\
2 & Qwen2.5-14B &  51.3 / 84.7 / 86.2 / 87.7        &   55.4 / 87.0 / 89.1 / 89.4      & {57.6} / {87.9} / {91.6} / {92.0}   &  57.6 / 67.3 / 79.1 / 81.3     &      60.2 / 70.6 / 82.1 / 84.9     &      {61.6} / {72.0} / {85.0} / {87.5} \\
3 & LLama-3-8B  &   40.9 / 77.5 / 80.4 / 78.1        &  54.4 / 86.6 / 88.1 / 88.2       & 56.1 / 87.6 / 90.3 / 91.1  &  59.1 / 67.4 / 81.4 / 85.0     &       60.1 / 70.6 / 82.6 / 85.2     &      61.6 / 72.0 / 85.0 / 87.8   \\
4 & Mistral-7B  &    38.1 / 76.4 / 79.8 / 77.6       &  53.6 / 86.3 / 87.7 / 88.1  & 54.7 / 87.2 / 90.1 / 90.5   &  46.8 / 60.1 / 67.1 / 72.2     &      59.7 / 70.2 / 81.7 / 84.0     &      61.3 / 71.9 / 82.2 / 85.4      \\ \midrule[1pt]
\multicolumn{8}{c}{\texttt{w/ Noisy Document}} \\ \midrule
5 & Qwen2.5-7B  &    48.6 / 83.4 / 84.6 / 86.1        &  54.7 / 86.7 / 88.5 / 89.1     & 56.5 / 87.7 / 91.1 / 91.8  & 55.2 / 66.0 / 74.0 / 76.9     &      59.0 / 69.7 / 81.2 / 83.0     &      61.1 / 71.9 / 84.2 / 86.1   \\
6 & Qwen2.5-14B &   49.4 / 83.7 / 84.9 / 86.5        &    55.5 / 87.0 / 89.2 / 89.4      & 57.7 / 87.9 / 91.6 / 92.1  & 56.3 / 66.2 / 77.8 / 80.2     &      60.2 / 70.6 / 81.9 / 84.6     &      61.6 / 72.2 / 85.1 / 87.8   \\
7 & LLama-3-8B  &   26.3 / 74.6 / 78.2 / 78.9        &   54.3 / 86.6 / 88.0 / 88.2     &   56.1 / 87.6 / 90.3 / 91.1   & 57.9 / 66.5 / 79.7 / 81.5     &      60.2 / 70.7 / 82.8 / 85.2     &      61.5 / 72.0 / 84.9 / 87.5       \\
8 & Mistral-7B  &    24.3 / 73.9 / 77.7 / 78.0       &   53.5 / 86.3 / 87.7 / 88.1      &    54.5 / 87.1 / 90.0 / 90.5   & 41.9 / 59.3 / 62.4 / 65.7     &      59.5 / 70.1 / 81.5 / 83.9     &      61.5 / 72.0 / 82.4 / 85.6       \\ \midrule[1pt]
\multicolumn{8}{c}{\texttt{w/ Golden English Document}} \\ \midrule
9 & Qwen2.5-7B  &     49.6 / 84.3 / 85.1 / 86.5       &   57.1 / 87.7 / 90.6 / 91.3       &   58.7 / 88.7 / 92.5 / 93.1     & 55.4 / 66.2 / 73.5 / 77.4     &      61.2 / 71.3 / 83.1 / 85.6     &      63.5 / 73.4 / 84.6 / 87.9    \\
10 & Qwen2.5-14B &     50.6 / 84.5 / 85.5 / 87.1       &   57.0 / 87.9 / 91.1 / 91.8     &  59.4 / 88.9 / 92.7 / 93.3     & 57.2 / 67.3 / 78.2 / 81.0     &      61.5 / 71.7 / 83.9 / 86.0     &      63.8 / 73.7 / 87.0 / 88.2     \\
11 & LLama-3-8B  &     29.4 / 75.3 / 80.2 / 80.9        &    55.8 / 87.6 / 89.7 / 90.4      &   58.0 / 88.6 / 91.4 / 92.0     & 58.7 / 67.2 / 80.9 / 85.3     &      61.7 / 71.7 / 84.0 / 86.5     &      64.2 / 73.6 / 87.8 / 89.1       \\
12 & Mistral-7B  &      26.6 / 74.8 / 79.8 / 80.3       &     54.9 / 87.3 / 89.5 / 89.8       &  56.6 / 88.1 / 91.1 / 91.9   & 42.3 / 60.2 / 66.7 / 69.6     &      61.4 / 71.5 / 83.5 / 85.8     &      63.4 / 72.9 / 85.8 / 87.9     \\ \midrule[1pt]
\multicolumn{8}{c}{\texttt{w/ Golden Chinese Document}} \\ \midrule
13 & Qwen2.5-7B  &     49.9 / 84.4 / 85.4 / 86.7        &   57.3 / 87.1 / 91.2 / 91.5     &  60.0 / 89.1 / 93.1 / 93.6   & 54.5 / 65.2 / 72.2 / 74.9     &      60.3 / 70.4 / 82.3 / 84.2     &      63.2 / 73.1 / 84.3 / 87.3       \\
14 & Qwen2.5-14B &    50.7 / 84.7 / 85.7 / 87.0        &    58.1 / 87.2 / 91.7 / 92.0     &   60.5 / 89.2 / 93.3 / 93.7   & 55.8 / 66.2 / 77.4 / 80.3     &      61.2 / 71.2 / 83.0 / 85.3     &      63.8 / 73.5 / 87.1 / 88.3       \\
15 & LLama-3-8B  &      35.7 / 76.4 / 80.8 / 81.0       &   57.1 / 87.0 / 90.2 / 90.8     &   59.5 / 89.0 / 92.8 / 93.2   & 57.0 / 66.7 / 79.3 / 83.1     &      61.1 / 71.0 / 82.7 / 84.9     &      63.8 / 73.2 / 87.5 / 89.3         \\
16 & Mistral-7B  &    34.9 / 75.2 / 79.8 / 80.4        &    56.4 / 87.8 / 89.9 / 90.3  &  58.8 / 88.8 / 92.6 / 93.0     & 41.8 / 59.0 / 65.8 / 68.2     &      60.7 / 70.8 / 82.5 / 84.6     &      63.3 / 73.1 / 85.9 / 88.2        \\ \midrule[1pt]
\multicolumn{8}{c}{\texttt{w/ Golden German Document}} \\ \midrule
17 & Qwen2.5-7B  &    44.2 / 83.2 / 84.3 / 85.1       &    55.6 / 87.2 / 88.9 / 89.6   &   58.7 / 88.7 / 92.4 / 93.1  & 55.8 / 66.4 / 76.9 / 77.8     &      62.0 / 72.3 / 83.6 / 85.2     &      64.5 / 74.5 / 85.9 / 90.0          \\
18 & Qwen2.5-14B &    45.2 / 82.9 / 84.8 / 85.6        &    56.3 / 87.5 / 89.4 / 89.7     &   59.4 / 88.9 / 92.7 / 93.1     & 57.4 / 67.5 / 78.6 / 80.9     &      62.6 / 72.7 / 84.2 / 85.7     &      65.1 / 74.7 / 88.2 / 90.4     \\
19 & LLama-3-8B  &     35.3 / 74.8 / 78.5 / 78.9        &      55.1 / 87.1 / 88.6 / 89.1     &    57.7 / 88.6 / 91.3 / 91.7    & 60.0 / 67.5 / 81.7 / 84.9     &      62.6 / 72.5 / 84.5 / 85.8     &      65.0 / 74.6 / 88.4 / 90.0      \\
20 & Mistral-7B  &      35.0 / 73.3 / 77.3 / 76.5        &      54.3 / 86.8 / 88.3 / 88.8     &   56.6 / 88.1 / 91.0 / 91.7     & 46.1 / 59.6 / 66.4 / 71.8     &      61.9 / 72.0 / 83.5 / 84.8     &      64.6 / 74.0 / 87.8 / 88.8    \\ \bottomrule[1pt]
\end{tabular}
}
\caption{Experimental results (BLEU / COMET / GRB / GRF) on RAGtrans. ``SFT LLMs'' denotes the LLMs are instruction-tuned on the training data of RAGtrans, while ``SFT+CSC LLMs'' denotes the LLMs are instruction-tuned on both RAGtrans and CSC multi-task training. \modi{All results of SFT+CSC LLMs are statistically significantly better than those of the corresponding SFT LLMs with t-test p < 0.01.}}
\label{table:main_res}
\end{table*}

\section{Experiments}

\subsection{Experimental Setup}

\noindent \textbf{Metrics.}
Following previous work, we adopt \emph{BLEU}~\cite{papineni-etal-2002-bleu} and reference-based \emph{COMET} score~\cite{rei-etal-2022-cometkiwi}.
BLEU evaluates n-grams overlap between the generated translations and corresponding references, while COMET evaluates the semantic similarity of translations against references.
Besides, recent studies~\cite{kocmi2023large,wang-etal-2023-chatgpt} also show the strong ability of LLMs in NLP evaluation.
Thus, we use evaluators implemented using GPT-4o in reference-based and reference-free styles, which we refer to as \emph{GRB} and \emph{GRF}, respectively.

\vspace{0.2ex}
\noindent \textbf{Backbones.}
We adopt four LLMs in the experiments: (1) Qwen2.5-7B-Instruct and (2) Qwen2.5-14B-Instruct~\cite{yang2024qwen2} are two cutting-edge Qwen-series LLMs.
(3) Llama-3-8B~\cite{dubey2024llama} is the latest llama-series LLM.
(4) Mistral-7B~\cite{jiang2023mistral} also shows great performance among the same-scale LLMs.

\vspace{0.2ex}
\noindent \textbf{Retriever.}
To support the full Wiki evaluation in RAGtrans, we implement two retrievers in the experiments:
(1) BM25~\cite{robertson2009probabilistic} is a traditional lexical search method that matches keywords efficiently with an inverted index. For a given source sentence, BM25 can retrieve its relevant documents only in the same language.
(2) BGE-m3~\cite{chen2024bge} is a multilingual sentence embedding model that supports dense retrievals across different languages.

\vspace{0.2ex}
\noindent \textbf{Implementation Details.}
For details about training hyper-parameters, SFT prompts, model checkpoints, training costs, CSC training samples and metric implementation, please refer to Appendix~\ref{appendix:implementation_details}.

\subsection{Main Results}

Table~\ref{table:main_res} shows the main results of the golden and robustness evaluation settings.
For each LLM, we evaluate its retrieval-augmented MT performance when giving empty, noisy or golden documents.

% \vspace{0.5ex}
\noindent \textbf{Zero-Shot Performance.}
\modi{Among all backbones, Qwen2.5-14B typically performs best in En$\Rightarrow$Zh while Llama-3-8B performs best in En$\Rightarrow$De.}
When giving noisy documents to LLMs, the MT performance of all LLMs decreases compared with those of giving empty documents.
For example, Qwen2.5-7B (w/ empty document) achieves 50.1 BLEU and 84.6 COMET in En$\Rightarrow$Zh, while the counterparts of Qwen2.5-7B (w/ noisy document) are 48.6 and 83.4.
This observation indicates the \emph{low robustness of zero-shot LLMs when faced with irrelevant documents}.
Moreover, when zero-shot LLMs use golden relevant documents as inputs, their MT performances do not increase (compared with those using empty documents) as expected.
Specifically, the performance either decreased or remained relatively stable (rows 9-20 vs. rows 1-4).
Besides, when given German documents in En$\Rightarrow$Zh translation, or given Chinese documents in En$\Rightarrow$De translation, the model performance decreases significantly.
Thus, \emph{the retrieval-augmented MT ability is limited in zero-shot LLMs, especially when the retrieved documents are in a language beyond the source and the target languages (named a third language)}.

\vspace{0.2ex}
\noindent \textbf{Instruction-Tuning Performance.}
After we tune LLMs on RAGtrans, their MT performance generally increases by a large margin.
For example, when giving empty documents in En$\Rightarrow$Zh, SFT Qwen2.5-7B outperforms zero-shot Qwen2.5-7B by 4.5 BLEU, 2.1 COMET, 3.0 GRB and 2.1 GRF.
\modi{Similarly, in En$\Rightarrow$De, SFT Qwen2.5-7B outperforms zero-shot Qwen2.5-7B by 3.1 BLEU, 3.0 COMET, 6.8 GRB and 5.3 GRF.}
In addition, we find that when using golden documents, SFT LLMs achieve better performance than those using empty documents, indicating that \emph{instruct-tuning on RAGtrans improves LLMs' retrieval-augmented MT ability}.
Even when giving the relevant documents in a third language, it can bring improvement.
\emph{The model robustness is also enhanced}, and the given noisy documents do not significantly perturb the model performance.
This is because a small number of training samples in RAGtrans consist of irrelevant documents as inputs, thus, LLMs can learn to translate source sentences conditioned on both relevant and irrelevant documents.

\vspace{0.2ex}
\noindent \textbf{CSC Training Performance.}
After instruction-tuning on RAGtrans and CSC multi-task training, the model performance is further improved.
\modi{When giving empty documents, CSC brings 1.1$\sim$2.2 BLEU and 0.9$\sim$1.0 COMET improvements compared with SFT LLMs in En$\Rightarrow$Zh, while bringing 1.4$\sim$1.7 BLEU and 1.4$\sim$2.1 COMET improvements compared with SFT LLMs in En$\Rightarrow$De (rows 1-4).}
This observation verifies the effectiveness of CSC, and \emph{LLMs' retrieval-augmented MT ability can be enhanced by the designed training objectives}.
\modi{Besides, taking Qwen2.5-14B as an example, when giving the relevant documents in a third language to SFT Qwen2.5-14B, it brings 0.9 BLEU and 0.5 COMET improvements in En$\Rightarrow$Zh  (row 18 vs. row 2) and 1.0 BLEU and 0.6 COMET in En$\Rightarrow$De (row 14 vs. row 2) compared with when giving empty documents. The counterpart improvements in SFT+CSC Qwen2.5-14B are 1.8 BLEU and 1.0 COMET in En$\Rightarrow$Zh, and 2.2 BLEU and 1.5 COMET in En$\Rightarrow$De.}
Therefore, \emph{CSC enhances LLMs' ability to leverage relevant knowledge in a third language}.
Moreover, we discuss the scalability of CSC in Appendix~\ref{appendix:scalability_of_csc}.

\begin{table}[t]
\centering
\resizebox{0.48\textwidth}{!}
{
\begin{tabular}{cclcc}
\toprule[1pt]
% \# & Knowledge                                                                      & \multicolumn{1}{c}{Method} & BLEU            & COMET         \\ \midrule[1pt]
\multirow{1}{*}{\#} & \multirow{1}{*}{Knowledge} & \multicolumn{1}{c}{\multirow{1}{*}{Method}} & \multicolumn{1}{c}{En$\Rightarrow$Zh} & \multicolumn{1}{c}{En$\Rightarrow$De} \\  \midrule[1pt]
1 & \multirow{2}{*}{\begin{tabular}[c]{@{}c@{}}Empty\\ Document\end{tabular}}  & Qwen2.5-7B         & 56.8 / 87.7            &   60.9 / 71.8         \\
2 &  & Qwen2.5-14B         &   57.6 / 87.9          &  61.6 / 72.0         \\ \midrule[1pt]
3 & \multirow{2}{*}{\begin{tabular}[c]{@{}c@{}}English\\ Wikipedia\end{tabular}}  & Qwen2.5-7B (+BM25)         & 57.4 / 88.0         &   61.9 / 72.2            \\
4 &  & Qwen2.5-14B (+BM25)         &   58.0 / 88.2      & 62.3 / 72.4             \\ \midrule[1pt]
5 & \multirow{2}{*}{\begin{tabular}[c]{@{}c@{}}English\\ Wikipedia\end{tabular}}  & Qwen2.5-7B (+BGEm3)       &     57.9 / 88.1         &  62.6 / 72.5         \\
6 & & Qwen2.5-14B (+BGEm3)         &  58.7 / 88.4     &  63.0 / 72.7           \\ \midrule[1pt]
7 & \multirow{2}{*}{\begin{tabular}[c]{@{}c@{}}Chinese\\ Wikipedia\end{tabular}}  & Qwen2.5-7B (+BGEm3)          &        58.3 / 88.3      &  62.4 / 72.5          \\
8 & & Qwen2.5-14B (+BGEm3)      &   59.0 / 88.5       & 62.8 / 72.6        \\ \midrule[1pt]
9 & \multirow{2}{*}{\begin{tabular}[c]{@{}c@{}}German\\ Wikipedia\end{tabular}}   & Qwen2.5-7B (+BGEm3)         &     57.9 / 88.1       &  63.2 / 73.1         \\
10 & & Qwen2.5-14B (+BGEm3)        &    58.7 / 88.3       & 63.6 / 73.4        \\ \midrule[1pt]
11 & \multirow{2}{*}{\begin{tabular}[c]{@{}c@{}}French\\ Wikipedia\end{tabular}}   & Qwen2.5-7B (+BGEm3)        &      57.6 / 88.0        &  62.5 / 72.4         \\
12 & & Qwen2.5-14B (+BGEm3)        &   58.3 / 88.3       & 62.7 / 72.6            \\ \midrule[1pt]
13 & \multirow{2}{*}{\begin{tabular}[c]{@{}c@{}}Czech\\ Wikipedia\end{tabular}}    & Qwen2.5-7B (+BGEm3)         &      57.3 / 87.9      &  61.8 / 72.2              \\
14 & & Qwen2.5-14B (+BGEm3)         &  57.9 / 88.1      & 62.4 / 72.5       \\ \midrule[1pt]
\end{tabular}
}
\caption{\modi{Experimental results of full wiki evaluation on SFT+CSC LLMs (BLEU / COMET).}}
\label{table:fullwikires}
\end{table}

\subsection{Full Wiki Evaluation}
Table~\ref{table:fullwikires} shows the experimental results on full Wiki evaluation.
We use the Wikipedia dumps in different languages as the knowledge sources for retrievers to retrieve relevant documents, and then leverage SFT+CSC LLMs to translate the source sentences (please refer to Appendix~\ref{appendix:details_of_full_wiki_testing} for more details).
Compared with using empty documents, retrieving documents from knowledge sources generally brings improvement, indicating the usability of retrieved knowledge.
Compared with the BM25 retriever, BGEm3 retriever helps LLMs achieve better performance (rows 5-6 vs. rows 3-4).
Besides, BGEm3 could retrieve knowledge from other languages, and the retrieved knowledge from a third language could also enhance model performance, verifying the SFT+CSC LLMs could leverage multilingual knowledge in retrieval-augmented MT.

\begin{table}[t]
\centering
\resizebox{0.45\textwidth}{!}
{
\begin{tabular}{lcccc}
\toprule[1pt]
  & \multicolumn{1}{c}{BLEU}   & \multicolumn{1}{c}{COMET}        & \multicolumn{1}{c}{BLEU}   & \multicolumn{1}{c}{COMET}         \\
 & \multicolumn{2}{c}{w/ Empty Doc.} & \multicolumn{2}{c}{w/ Golden En.} \\ \midrule[1pt]
Qwen2.5-7B (SFT+CSC)      & 56.8            & 87.7          & 58.7            & 88.7          \\
$\quad$- CLIC                    & 56.5             &  87.6             &   57.9      &  88.3          \\
$\quad$- SKET       &  56.1      &  87.5      &  57.5     &  88.1   \\
$\quad$- CLRD       &   56.5    &   87.7    &  58.3     &   88.5      \\ \midrule[1pt]
                          & \multicolumn{2}{c}{w/ Golden Zh.} & \multicolumn{2}{c}{w/ Golden De.} \\ \midrule[1pt]
Qwen2.5-7B (SFT+CSC)      & 60.0            & 89.1          & 58.7            & 88.7          \\
$\quad$- CLIC             &   59.5    &  88.9     &  57.0     &    88.2       \\
$\quad$- SKET              &   58.7    &  88.5     &     57.9   &   88.5          \\
$\quad$- CLRD                 &  59.7     &  89.0     &   56.7    &   87.9  \\ \bottomrule[1pt]            
\end{tabular}
}
\caption{Ablation study on En$\Rightarrow$Zh golden evaluation. Doc.: Document; ``En.'', ``Zh.'' and ``De.'' indicate English, Chinese and German documents, respectively.} 
\label{table:ablations}
\end{table}

\subsection{Ablations}
\label{subsec:ablations}
As shown in Table~\ref{table:ablations}, we conduct ablation studies on En$\Rightarrow$Zh golden evaluation\footnote{\modi{For ablations on En$\Rightarrow$De golden evaluation, please refer to Appendix~\ref{appendix:ablation_en2de}, which show similar trends to En$\Rightarrow$Zh.}} to figure out the contributions of each training objective in CSC.
Specifically, we remove each objective, and evaluate the model performance accordingly.
In each case, the performance is lower than using all training objectives, indicating the effectiveness of every objective.
When giving empty documents or documents in source/target language, the most important objective is self-knowledge-enhanced translation (SKET) since it enhances models' MT ability.
When giving documents in a third language, cross-lingual information completion (CLIC) and cross-lingual relevance discrimination (CLRD) become more important than SKET, since these two objectives train LLMs to refine and judge information from multilingual knowledge.

\subsection{Human Evaluation}

\begin{table}[t]
\centering
\resizebox{0.48\textwidth}{!}
{
\begin{tabular}{clc|ccccc}
\toprule[1pt]
& \multicolumn{2}{c|}{Method}                 & \multicolumn{5}{c}{Error Type (\%)}                          \\
\# & \multicolumn{1}{c}{Model} & Document       & Ref. & Word & Phrase & Fluency & Other \\ \midrule[1pt]
1 & Qwen2.5-7B (SFT)          & Empty          & 0.50      & 7.00       & 3.33         & 1.50    & 0.50  \\
2 & Qwen2.5-14B (SFT)         & Empty          & 0.33      & 5.50       & 2.67         & 1.33    & 0.50  \\
3 & Qwen2.5-7B (SFT+CSC)      & Empty          & 0.50      & 5.50       & 2.67         & 1.50    & 0.50  \\
4 & Qwen2.5-14B (SFT+CSC)     & Empty          & 0.50      & 4.67       & 2.33         & 1.00    & 0.17  \\
5 & Qwen2.5-7B (SFT)          & Golden Zh. & 0.33      & 5.17       & 2.50         & 1.33    & 0.50  \\
6 & Qwen2.5-14B (SFT)         & Golden Zh. & 0.17      & 4.67       & 2.33         & 1.33    & 0.00  \\
7 & Qwen2.5-7B (SFT+CSC)      & Golden Zh. & 0.17      & 5.00       & 2.17         & 1.50    & 0.33  \\
8 & Qwen2.5-14B (SFT+CSC)     & Golden Zh. & 0.17      & 3.50       & 1.50         & 1.33    & 0.33  \\ \bottomrule[1pt]
\end{tabular}
}
\caption{Human evaluation (En$\Rightarrow$Zh) of the retrieval-augmented MT ability. Ref.: Reference.} 
\label{table:human_evaluation_ramta}
\end{table}

\noindent \textbf{Retrieval-Augmented MT Ability.}
We employ human evaluation to further study the MT performance of SFT LLMs and SFT+CSC LLMs.
Specifically, human evaluators judge whether the translations include the following flaws: reference errors, word-level errors, phrase-level errors, fluency flaws and other errors (more details are given in Appendix~\ref{appendix:human_evaluation}).
As shown in Table~\ref{table:human_evaluation_ramta}, the two most common error types are word-level and phrase-level errors.
In these two types, the CSC multi-task training method and the golden documents enhance the model performance, verifying the effectiveness of CSC and the usability of the golden documents.

\vspace{0.5ex}
\noindent \textbf{The Effects of Documents.}
To study the effects of documents, we provide the different documents to Qwen2.5-7B (SFT+CSC), including empty, noisy, golden English, golden Chinese and golden German documents, to evaluate if the corresponding translations involve flaws.
As shown in Table~\ref{table:human_evaluation_teod}, the noisy documents still increase the number of translation flaws (row 2 vs. row 1).
Robustness is a crucial factor for the deployment of LLMs in real applications, thus \emph{future work could pay more attention to model robustness}.
Besides, we also find that when providing the model with golden Chinese or English documents, the number of translation flaws typically decreases. However, when providing golden German documents, the number of word-level errors significantly increases (row 5 vs. row 1).
We further observe the cases, and find this is because \emph{the German documents might encourage MT models to translate some entities to German if the entities listed in the documents, thus raising word-level errors}.
This issue should also be noticed in future work, since the retrieval-argument MT models might receive documents from various languages in real applications.

\begin{table}[t]
\centering
\resizebox{0.45\textwidth}{!}
{
\begin{tabular}{clccccc}
\toprule[1pt]
\multicolumn{1}{c}{\multirow{2}{*}{\#}} &\multicolumn{1}{c}{\multirow{2}{*}{Document}} & \multicolumn{5}{c}{Error Type (\%)}                     \\
& \multicolumn{1}{c}{}                          & Ref. & Word & Phrase & Fluency & Other \\ \midrule[1pt]
1 & Empty                                         & 0.50      & 5.50       & 2.67         & 1.50    & 0.50  \\
2 & Noisy                                         & 1.83      & 6.67       & 5.33         & 1.83    & 1.50  \\
3 & Golden Chinese                                & 0.17      & 5.00       & 2.17         & 1.50    & 0.33  \\
4 & Golden English                                & 0.17      & 3.17       & 1.33         & 0.83    & 0.17  \\
5 & Golden German                                 & 0.50      & 8.17       & 1.67         & 1.33    & 0.33  \\ \bottomrule[1pt]
\end{tabular}
}
\caption{Human evaluation  (En$\Rightarrow$Zh) of the effects of documents, using Qwen2.5-7B (SFT+CSC) as the MT model.} 
\label{table:human_evaluation_teod}
\end{table}

\section{Related Work}

To leverage additional knowledge to enhance MT performance, previous literature typically explores paired sentences (also known as ``translation memory'') or structured knowledge graphs as the knowledge sources:
(1) \emph{Paired Sentences}:
\citet{zhang-etal-2018-guiding} utilize a search engine to retrieve sentence pairs whose source sides are similar to the input sentences.
\citet{bulte-tezcan-2019-neural} design a fuzzy retriever to enhance the model performance.
\citet{he-etal-2021-fast} design a fast and accurate method to improve the robustness of pair-sentence-enhanced MT models.
\citet{cai-etal-2021-neural} relax the bilingualism limitation in retrieving paired sentences, and they try to retrieve similar target-language sentences to enhance MT models.
(2) \emph{Knowledge Graphs}: A few studies leverage relevant information from structured knowledge graphs to enhance MT models.
\citet{conia-etal-2024-towards} use Wikidata~\cite{vrandevcic2014wikidata}, a multilingual knowledge graphs, to enhance MT models.
\citet{chen2024crat} build an internal knowledge graph based on context, and then use it to enhance translation.
Different from previous work, we aim to utilize unstructured documents to provide knowledge to MT models.
% Furthermore, these documents can be in various languages and do not require any alignment across languages.

\section{Conclusion}

In this paper, we explore the retrieval-augmented MT with unstructured knowledge.
To this end, we build RAGtrans dataset with 169K retrieval-augment MT samples to train and evaluate LLMs' retrieval-augmented MT ability.
Further, we propose CSC multi-task training method with three designed objectives to teach LLMs to leverage multilingual knowledge in retrieval-augmented MT.
Extensive experiments demonstrate the usability of RAGtrans and the effectiveness of CSC.

\section*{Limitations}

While we show LLMs' retrieval-augmented MT ability and the effectiveness of CSC multi-task training method, there are some limitations worth noting:
(1) For multilingual knowledge bases, we use Wikipedia in some specific languages (\emph{e.g.}, Chinese, English, German, French, and Czech). Future work could extend the multilingual sources to other languages or other sources.
(2) During data collection of RAGtrans, a CoT prompt is used in the GPT-4o translation (c.f. Figure~\ref{fig:gpt_4o_translation}). However, in the SFT process, we do not use the CoT prompt to train LLMs, and future work could explore the effect of CoT in retrieval-augmented MT.

\section*{Ethical Considerations}

We discuss the main ethical considerations of RAGtrans as follows:
(1) \emph{Licenses}. The source sentences and documents are derived from Wikipedia, whose texts are under CC BY-SA and GFDL licenses.
We will release the RAGtrans dataset under CC-BY-SA 4.0 license.
(2) \emph{Compensation}. During the translation annotation, the salary for translating each sentence is determined by the average time of annotation and local labor compensation standard.

\bibliography{custom}

\appendix

\begin{figure*}[t]
\centerline{\includegraphics[width=0.96\textwidth]{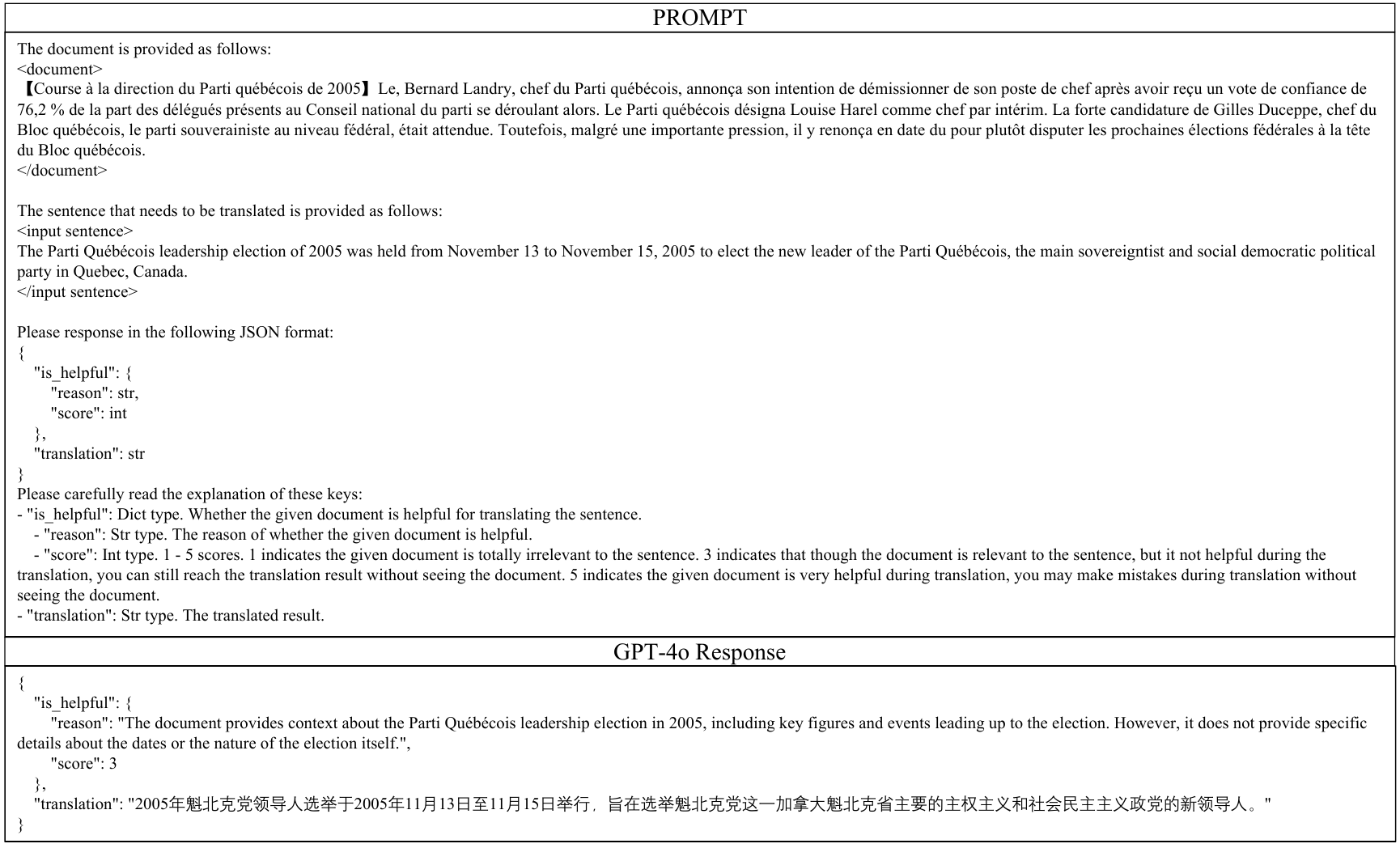}}
\caption{An En$\Rightarrow$Zh example of the complete prompt in GPT-4o translator.}
\label{fig:gpt_4o_translation_complete_prompt}
\end{figure*}

\section{Details of GPT-4o translation}
\label{appendix:gpt4o_trans}

\noindent \textbf{Complete Prompt.}
We provide the system prompt as follows: ``\emph{You are a professional translator, and your task is to translate a given input sentence from English to Chinese/German. In addition to the input sentence, you will be provided with a document that may contain relevant information to aid in the translation. However, be aware that some documents may contain irrelevant or noisy information}''.
An En$\Rightarrow$Zh example of user prompt and model response is shown in Figure~\ref{fig:gpt_4o_translation_complete_prompt}, where both a (French) document and an English source sentence are provided in the user prompt.
We also define a 5-point rating breakdown to align the scoring value between GPT-4o and humans.
In the model response, GPT-4o first judges the relevance between the given document and the source sentence, and then provides the corresponding translation.

\vspace{0.5ex}
\noindent \textbf{Quality Analysis.}
To figure out the quality of GPT-4o translations, we calculate the reference-free CometKiwi score between the source English sentences and GPT-4o translations. As a result, the average score of En$\Rightarrow$Zh translation is 84.48, \modi{and En$\Rightarrow$De translation is 84.90}, indicating high translation quality~\cite{rei-etal-2022-cometkiwi}.

\vspace{0.5ex}
\noindent \textbf{Other Details.}
The version of GPT-4o used in this work is \emph{GPT-4o-2024-08-06}.
When calling the official APIs, we set the temperature to 0.1, and set default values for other hyper-parameters.

\begin{figure*}[t]
\centerline{\includegraphics[width=0.95\textwidth]{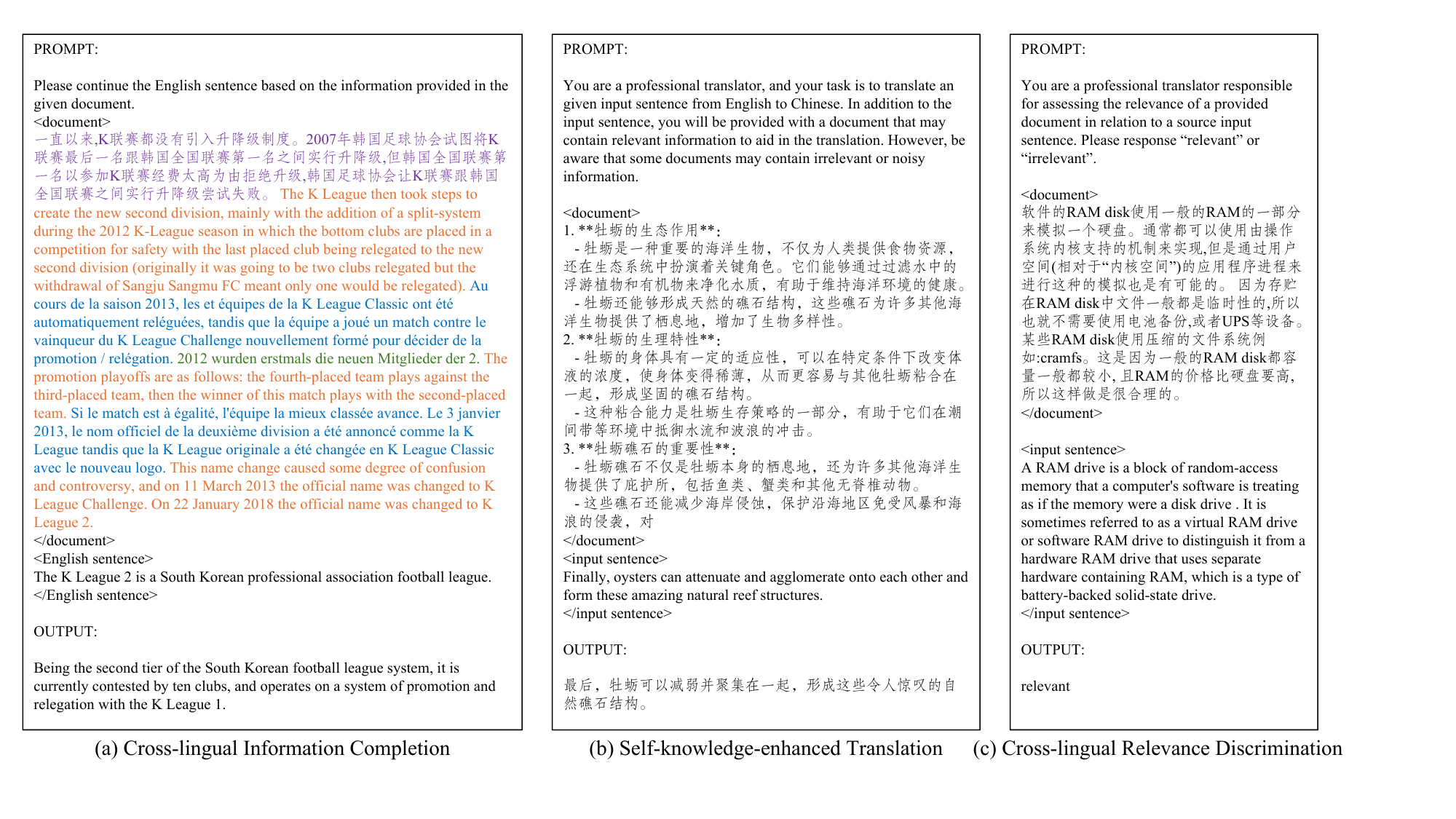}}
\caption{Examples of CSC training objectives. Different colors in (a) means different langauges, including \textcolor[RGB]{112,48,160}{Chinese}, \textcolor[RGB]{233,113,50}{English}, \textcolor[RGB]{0,112,192}{French} and \textcolor[RGB]{59,125,35}{German}.}
\label{fig:CSC_examples}
\end{figure*}

\section{Implementation Details}
\label{appendix:implementation_details}

\noindent \textbf{Training Details.}
Llama-Factory~\cite{zheng-etal-2024-llamafactory} is used to instruct-tune LLMs.
All LLMs are tuned on 8$\times$NVIDIA A100 GPUs (40G) with 1e-5 learning rate and 32 (8$\times$4) batch size.
We use the DeepSpeed optimization~\cite{rasley2020deepspeed}, and set ZeRO-2 optimization for Qwen2.5-7B-Instruct and Mistral-7B, while ZeRo-3 for Qwen2.5-14B-Instruct and Llama-3-8B.
During tuning, documents are also truncated to ensure the input length is within 2K tokens.

\vspace{0.5ex}
\noindent \textbf{SFT prompt.}
The system prompt in SFT is the same as the GPT-4o translator (c.f. Appendix~\ref{appendix:gpt4o_trans}).
The user prompt in SFT is provided as follows: ``\texttt{<document>}[doc]\texttt{</document>}\texttt{<input sentence>}[sent]\texttt{</input sentence>}'', where \texttt{<document>}, \texttt{</document>}, \texttt{<input sentence>} and \texttt{</input sentence>} are special tokens to indicate the boundaries of the given document (denoted as ``[doc]'') and the source sentence (``[sent]'').

\vspace{0.5ex}
\noindent \textbf{Model Checkpoints.}
We use four LLM backbones in experiments, \emph{i.e.}, Qwen2.5-7B-Instruct\footnote{\url{https://huggingface.co/Qwen/Qwen2.5-7B-Instruct}}, Qwen2.5-14B-Instruct\footnote{\url{https://huggingface.co/Qwen/Qwen2.5-14B-Instruct}}, Llama-3-8B\footnote{\url{https://huggingface.co/hfl/llama-3-chinese-8b-instruct-v3}} and Mistral-7B\footnote{\url{https://huggingface.co/mistralai/Mistral-7B-Instruct-v0.3}}.
All model checkpoints are available at Huggingface.co community.

\vspace{0.5ex}
\noindent \textbf{Training Hours.}
All experiments are conducted on NVIDIA A100 GPUs with 40G memory, and we use its GPU hours to denote the consumption of computing resources.
We SFT LLMs on the training data of RAGtrans with 2 epochs, and each epoch costs 9.1 GPU hours, 54.0 GPU hours, 33.5 GPU hours, and 9.3 GPU hours for Qwen2.5-7B-Instruct, Qwen2.5-14B-Instruct, Llama-3-8B and Mistral-7B, respectively.
For SFT+CSC LLMs, more GPU hours are costed.
For example, to SFT Qwen2.5-14B-Instruct on both the RAGtrans training samples and CSC samples, each epoch costs 208 GPU hours; while the counterparts of Qwen2.5-7B-Instruct, Llama-3-8B and Mistral-7B are 34.1, 128.0 and 33.9 GPU hours, respectively.

\vspace{0.5ex}
\noindent \textbf{Multi-Task Training Samples.}
As we introduce in Section~\ref{sec:3}, there are three training objectives in CSC multi-task training method.
To provide a deeper understanding of these objectives, here we give some example samples in Figure~\ref{fig:CSC_examples}.
In our main experiments, we create 40K samples for each training objective.

\begin{figure}[t]
\centerline{\includegraphics[width=0.48\textwidth]{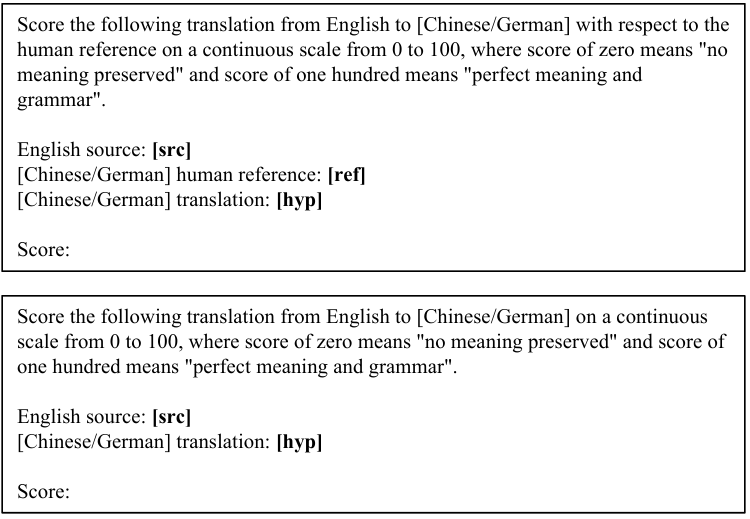}}
\caption{The prompts in GRB (upper part) and GRF (lower part). ``[src]'', ``[ref]'' and ``[hyp]'' denote the source sentence, human translation and model translation, respectively. ``[Chinese/German]'' denotes ``Chinese'' or ``German'', depending on the target language.}
\label{fig:gpt_4o_evaluator}
\end{figure}

\vspace{0.5ex}
\noindent \textbf{Metric Implementation.}
To calculate the reference-based COMET score~\cite{rei-etal-2022-cometkiwi}, we leverage the official codes\footnote{\url{https://github.com/Unbabel/COMET}} and the official model\footnote{\url{https://huggingface.co/Unbabel/wmt22-comet-da}}.
To calculate the BLEU score, we use the \textit{sacrebleu} toolkit\footnote{\url{https://github.com/mjpost/sacrebleu}} to calulate the corpus-level BLEU.
\modi{In En$\Rightarrow$Zh and En$\Rightarrow$De translation, the ``tokenize'' is set to ``zh'' and ``char'', respectively.}
For GRB and GRF, we prompt GPT-4o (2024-08-06 version) as the MT evaluator in the reference-based and reference-free manners, respectively.
The corresponding prompts borrow from \citet{kocmi2023large}, and are illustrated in Figure~\ref{fig:gpt_4o_evaluator}.
Since GRB and GRF need the API costs, we randomly select 200 samples from the RAGtrans testing set, and conduct the GRB and GRF evaluation.
All experimental results listed in this paper are the average of 3 runs.

\begin{figure*}[t]
\centering
\subfigure{
  \includegraphics[width=0.48\linewidth]{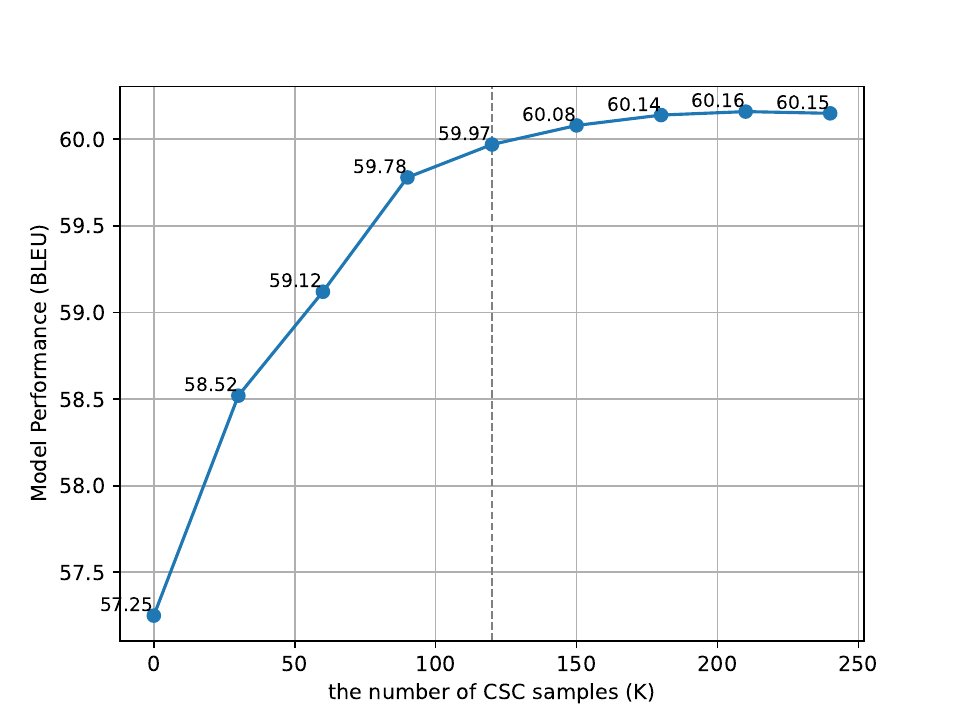}
}
\subfigure{
  \includegraphics[width=0.48\linewidth]{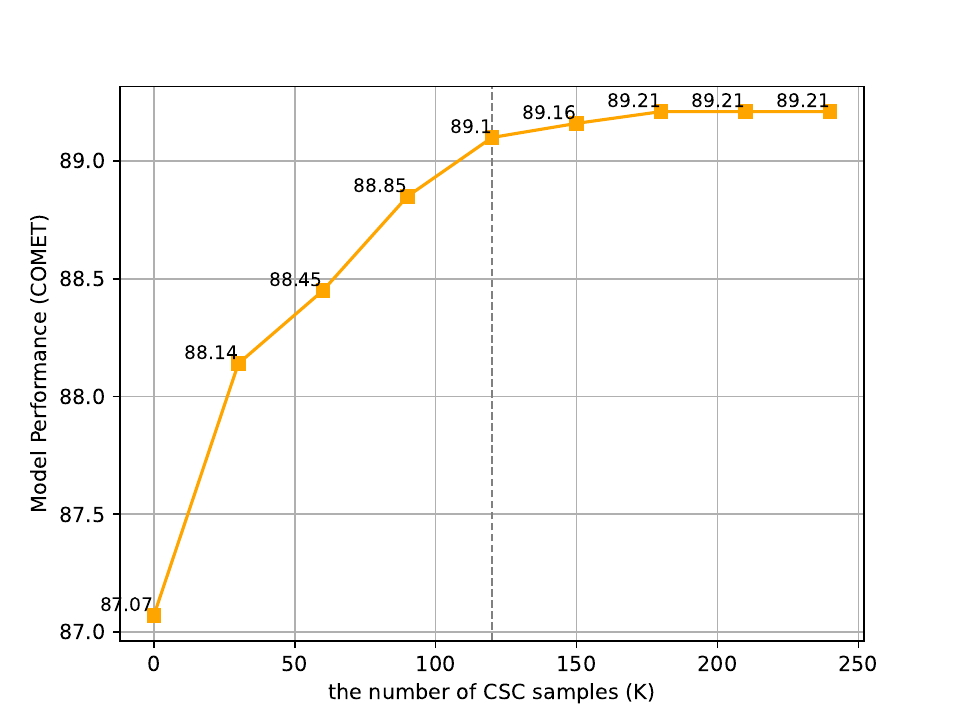}
}
\caption{The experimental results of CSC scalability (in En$\Rightarrow$Zh).}
\label{fig:csc_scalability}
\end{figure*}

\section{Scalability of CSC}
\label{appendix:scalability_of_csc}

As we demonstrate the effectiveness of CSC multi-task training method in experiments, we wonder the upper limit of the improvement brought by CSC.
To this end, we use Qwen2.5-7B-Instruct as the backbone, and systematically vary the number of CSC samples during the instruction tuning to examine the resulting performance (w/ golden Chinese document) changes.
As shown in Figure~\ref{fig:csc_scalability}, when the number of CSC samples exceeds 120K, the improvement brought by CSC begins to plateau.
When the number of CSC samples increases from 210K to 240K, the model performance does not improve accordingly.

\section{Details of Full Wiki Testing}
\label{appendix:details_of_full_wiki_testing}

\noindent \textbf{Retriever.}
For BM25 retriever, we use the implementation of \textit{elasticsearch}\footnote{\url{https://github.com/elastic/elasticsearch}} toolkit to retrieve top-$3$ documents for each source sentence.
For BGE-m3 retriever, we first use BGE-m3 sentence embedding model\footnote{\url{https://huggingface.co/BAAI/bge-m3}} to calculate the embedding of all documents in knowledge sources, and then use the embedding of source sentence to retrieve top-$3$ relevant documents via FAISS~\cite{johnson2019billion}.

\vspace{0.5ex}
\noindent \textbf{Knowledge Sources.}
We use Wikipedia dumps (20241001 version) as the knowledge sources, and leverage \textit{wikiextractor}\footnote{\url{https://github.com/attardi/wikiextractor}} toolkit to extract articles from Wikipedia dumps.
Following \citet{karpukhin2020dense}, we split each article into multiple, disjoint text blocks of 100 words as passages, serving as our basic retrieval units.
In the full Wiki evaluation, we build the knowledge sources based on English, Chinese, German, French, Czech, Russian, Korean and Japanese Wikipedia dumps, resulting in tens of millions of retrieval units.

\begin{table}[t]
\centering
\resizebox{0.40\textwidth}{!}
{
\begin{tabular}{lcccc}
\toprule[1pt]
  & \multicolumn{1}{c}{BLEU}   & \multicolumn{1}{c}{COMET}        & \multicolumn{1}{c}{BLEU}   & \multicolumn{1}{c}{COMET}         \\
 & \multicolumn{2}{c}{w/ Empty Doc.} & \multicolumn{2}{c}{w/ Golden En.} \\ \midrule[1pt]
Qwen2.5-7B (SFT+CSC)      & 60.9            & 71.8          & 63.5            & 73.4          \\
$\quad$- CLIC                    &   60.4           &     71.5          &   62.8     &    72.9        \\
$\quad$- SKET       &   59.8    &   71.0    &   62.2    &   72.3  \\
$\quad$- CLRD       &   60.5   &   71.5    &  63.0     &  73.1      \\ \midrule[1pt]
                          & \multicolumn{2}{c}{w/ Golden Zh.} & \multicolumn{2}{c}{w/ Golden De.} \\ \midrule[1pt]
Qwen2.5-7B (SFT+CSC)      & 63.2            & 73.1          & 64.5            & 74.5          \\
$\quad$- CLIC             &  62.4   &   72.6     &    63.4   &   73.7       \\
$\quad$- SKET              &   62.9  &  72.9     &   62.9    &  73.2          \\
$\quad$- CLRD                 &    62.2   &  72.5     &    63.9   &   73.9   \\ \bottomrule[1pt]            
\end{tabular}
}
\caption{\modi{Ablation study on En$\Rightarrow$De golden evaluation. Doc.: Document; ``En.'', ``Zh.'' and ``De.'' indicate English, Chinese and German documents, respectively.}} 
\label{table:ablations_en2de}
\end{table}

\section{\modi{Ablations on En$\Rightarrow$De}}
\label{appendix:ablation_en2de}

\modi{Table~\ref{table:ablations_en2de} shows the ablation results on En$\Rightarrow$De golden evaluation. The results also demonstrate that SKET is more important than the other two tasks when given empty documents or documents in the source/target language.
When given documents in a third language, the importance of SKET decreases and is less than that of CLIC/CLRD.
The conclusion is consistent with that in Section~\ref{subsec:ablations}.}

\section{Details of Human Evaluation}
\label{appendix:human_evaluation}

\noindent \textbf{Evaluators.}
Three master students are recruited in our human evaluation, and they are fluent in both Chinese and English.

\vspace{0.5ex}
\noindent \textbf{Instruction.}
The human evaluators are provided with the instructions for each translation error type:
(1) Reference Errors: Are there any mistakes in pronoun or reference usage that could cause confusion about what or whom is being referred to?
(2) Word-Level Errors: Are there incorrect translations, omissions, or additions of individual words that alter the meaning of the text?
(3) Phrase-Level Errors: Are there incorrect translations, omissions, or additions of phrases that affect the overall coherence and accuracy of the translation?
(4) Fluency Issues: Does the translation flow smoothly, or are there awkward phrases or constructions that impede comprehension?
(5) Other Errors: Are there any additional errors present in the translation that do not fit into the categories above?

\vspace{0.5ex}
\noindent \textbf{Evaluation Samples.}
Since human evaluation is labor-intensive, we randomly select 200 samples from the En$\Rightarrow$Zh testing set of RAGtrans to conduct the human evaluation.

\vspace{0.5ex}
\noindent \textbf{Inter-agreement.}
The Fleiss' Kappa scores~\cite{fleiss1971measuring} of the five error types are 0.63, 0.57, 0.68, 0.75 and 0.66 in our human evaluation, respectively, indicating a good inter-agreement among our evaluators.

\vspace{0.5ex}
\noindent \textbf{Other Details.}
\modi{During our human evaluation, we find that a golden Chinese document might involve multiple domain terms that are similar in lexical but with slight discrepancies. During translation, LLMs might use a flawed Chinese term (similar to the right term from the golden Chinese document) in the translation results. This will make them achieve good results in terms of automatic evaluation (including GRF and GRB). However, in our human evaluation, we will judge them as translation errors.}

\begin{table}[t]
\centering
\resizebox{0.45\textwidth}{!}
{
\begin{tabular}{lcccc}
\toprule[1pt]
\multicolumn{1}{c}{SFT LLMs}                          & BLEU & COMET & GRB  & GRF  \\ \midrule[1pt]
Qwen2.5-7B (w/o LP.)  & 57.3 & 87.1  & 91.2 & 91.5 \\
Qwen2.5-14B (w/o LP.) & 58.1 & 87.2  & 91.7 & 92.0   \\
Llama-3-8B (w/o LP.)  & 57.1 & 87.0    & 90.2 & 90.8 \\
Mistral-7B (w/o LP.)  & 56.4 & 87.8  & 89.9 & 90.3 \\ \midrule[1pt]
Qwen2.5-7B (w/ LP.)   & 57.9 & 87.7  & 92.1 & 92.3 \\
Qwen2.5-14B (w/ LP.)  & 58.6 & 87.9  & 92.4 & 92.5 \\
Llama-3-8B (w/ LP.)   & 57.7 & 87.5  & 90.9 & 91.4 \\
Mistral-7B (w/ LP.)   & 57.1 & 88.0    & 90.9 & 91.1 \\ \bottomrule[1pt]
\end{tabular}
}
\caption{\modi{Experimental results (En$\Rightarrow$Zh) with and without lead paragraphs of golden Chinese documents. w/: with; w/o: without; LP.: lead paragraphs.}} 
\label{table:exp_w_leadpara}
\end{table}

\section{Experiments with Lead Paragraphs}

\modi{As illustrated in Section~\ref{sec:ragtrans}, When constructing documents, we remove the lead paragraphs (\emph{i.e.}, $D^{\textit{l}} \setminus p^{\textit{l}}_1 (\textit{l} \in \{\text{en}, \text{zh}, \text{de}, \text{fr}, \text{cs}\})$).
This is under the considerations of (a) answer leakage and (b) consistency with real scenes. For (a), the answer leakage issue will exist if we do not remove the lead paragraphs for the target-language documents. This is because the lead paragraphs in the target-language documents might involve the translation fragments of source sentences. For (b), in real-world applications, for a source sentence, if we can retrieve a document that involves its translation or translation fragments, it will be useful for translating the sentence. But in most cases, we cannot guarantee that the retrieved documents involve translations or translation fragments due to many practical difficulties such as insufficient knowledge corpus, limited ability in cross-lingual retriever, etc. Therefore, we also remove the lead paragraphs from documents in other languages (beyond the target language).}

\modi{To figure out the effect of lead paragraphs in non-English documents, we conduct a new experiment to evaluate model performance equipped with them.
Specifically, we add the lead paragraphs of the Chinese documents for each En$\Rightarrow$Zh test sample, and re-evaluate their performance (w/ Golden Chinese document) in the test set. As shown in Table~\ref{table:exp_w_leadpara}, with the help of lead paragraphs, the model performance is generally improved.}

\end{document}